\documentclass[letterpaper, 10 pt, journal, twoside]{ieeeconf}  % Comment this line out if you need a4paper
\IEEEoverridecommandlockouts                              % This command is only needed if you want to use the \thanks command
\usepackage{capt-of,graphicx,subcaption}
\usepackage{xargs}                      % Use more than one optional parameter in a new commands
\usepackage[pdftex,dvipsnames]{xcolor}  % Coloured text etc.
\usepackage[colorinlistoftodos,prependcaption]{todonotes}
\usepackage{amsmath,amssymb,bm,mathtools}
\usepackage{cite}
\usepackage[font={small,it}]{caption}
\usepackage{relsize}
\usepackage[utf8]{inputenc}
\usepackage[T1]{fontenc}
\usepackage{textcomp}
\usepackage{gensymb} %% for the degree symbol
\usepackage{tikz}
\usetikzlibrary{shadings}
\usepackage[margin=1.5cm]{geometry} %% Modify page margins if needed
\usepackage{algorithm}
\usepackage[noend]{algpseudocode}
\usepackage[subpreambles]{standalone} %% include standalone pseudocodes
\usepackage{hyperref}
\makeatletter
\def\BState{\State\hskip-\ALG@thistlm}
\makeatother

\newcommandx{\oliver}[2][1=]{\todo[linecolor=Orchid,backgroundcolor=Goldenrod!25,bordercolor=Orchid,#1,inline]{Oliver: #2}}

\newcommand{\Project}{LeagueAI}

\graphicspath{{./sections/figures/}} % Define path to folder containing figures

\title{\LARGE \textbf{\Project: Improving object detector performance and flexibility through automatically generated training data and domain randomization}}

\author{Oliver Struckmeier
\thanks{Oliver Struckmeier, is with the Department of Electrical Engineering and Automation, Aalto University, Espoo 02150, Finland \tt \footnotesize \{oliver.struckmeier\}@aalto.fi}
}

\begin{document}
\maketitle
\begin{abstract}
In this technical report I present my method for automatic synthetic dataset generation for object detection and demonstrate it on the video game \textit{League of Legends}.
This report furthermore serves as a handbook on how to automatically generate datasets and as an introduction on the dataset generation part of the LeagueAI framework.

The LeagueAI framework is a software framework that provides detailed information about the game League of Legends based on the same input a human player would have, namely vision.
The framework allows researchers and enthusiasts to develop their own intelligent agents or to extract detailed information about the state of the game.

A big problem of machine vision applications usually is the laborious work of gathering large amounts of hand labeled data.
Thus, a crucial part of the vision pipeline of the LeagueAI framework, the dataset generation, is presented in this report.
The method involves extracting image raw data from the game's 3D models and combining them with the game background to create game-like synthetic images and to generate the corresponding labels automatically.

In an experiment I compared a model trained on synthetic data to a model trained on hand labeled data and a model trained on a combined dataset.
The model trained on the synthetic data showed higher detection precision on more classes and more reliable tracking performance of the player character.
The model trained on the combined dataset did not perform better because of the different formats of the older hand labeled dataset and the synthetic data.
\end{abstract}
%% abstract

%% Introduction
\section{Introduction and related Work}
In the recent years Deep Neural Networks and in particular Convolutional Neural Networks (CNNs) have been established as state of the art techniques for image classification and object detection ~\cite{szegedy2015going,rajpura2017object}.
Current state of the art object classification methods such as \textit{Faster R-CNN} and \textit{SSD} are widely used and provide good results in real-time object classification.
Another approach is \textit{You Only look Once} (YOLO), as presented in \cite{redmon2016you}, in which a single neural network predicts object classes and bounding box positions in one detection iteration.
In the latest iteration, \textit{YOLOv3}~\cite{redmon2018yolov3}, Redmon and Farhadi showed that their method provides a significant speed improvement compared to SSD and Faster R-CNN.
This speed-up allows the application of object detection to real-time applications.

However, training neural networks for object classification tasks requires vast amounts of training data, which is usually difficult to obtain because it requires human generated annotations.
Labeling images is a laborious and time intensive task and is a major obstacle when designing new applications based on object detection and classification.
In recent research domain randomization is used in simulation for training DNNs and then transferring them to real world applications~\cite{tobin2017domain}.
Further research has investigated how synthetic images can be used to train CNNs~\cite{rajpura2017object,prakash2018structured}.
Rajpura et al. found that synthetic images created from rendering 3D-models in a randomized 3D-scene provide less precise object detection, but when combined with real data lead to a higher mean average precision (mAP).
In a recent work Prakash et al. proposed a new method for synthetic dataset generation that takes the structure and the context of a scene into account.
Trained to detect cars using only synthetic data and benchmarked on the KITTI benchmark suite, the method outperformed other state of the art synthetic data generation methods.
Furthermorel, the paper supports the claim that combining synthetic and hand labeled data further improves the performance of the trained model.

It is evident that the capability to automatically generate training data for new domains can drastically reduce the effort required to train new models and allow for more flexible and efficient applications using object detection.

In this paper I present a method to generate large synthetic randomized datasets with precise labels by generating two dimensional masked raw data from 3D models of objects, combining them with backgrounds from the game to generate training data.
Finally different levels of noise and randomization are applied to the images to make the training data more variable and thus the model more robust.
The dataset generation pipeline and the benefits of large and randomized datasets versus hand labeled datasets are demonstrated by training different \textit{YOLOv3} object classifiers for the video game \textit{League of Legends} and comparing their mean average precision (mAP).
Furthermore the detection precision of the trained model is compared to a model trained on hand labeled data and to a model trained with both hand labeled and synthetic data.

The results show that a model trained with the synthetic data is capable of achieving significantly higher mAP with more detectable classes while taking only a fraction of the time it takes to label the images by hand.
Combining synthetic and hand labeled data did not further improve mAP as a result of differing image aspect ratios of the older hand labeled dataset.
The detection speed allow real-time use of the detectors at roughly 12-18 frames per second on a Nvidia GTX1080 GPU while running the game on the same machine.

\section{Code}
In this section the structure of the code of this project is briefly explained.
All snapshot of the currently private repository is available on \href{https://github.com/Oleffa/LeagueAI}{GitHub}\footnote{https://github.com/Oleffa/LeagueAI}.
\begin{itemize}
\item \texttt{LeagueAI\_helper.py}: The core of the framework providing all important functions for object detection, loading datasets etc.
\item \texttt{LeaugeAI\_minimal\_example.py}: A minimal example demonstrating how to use the framework with different inputs such as videos, images and in real-time on the game.
\item \texttt{LeagueAI\_vayne\_bot.py}: An implementation of the bot from the old project with the new LeagueAI framework.
\item \texttt{LeagueAI\_mAP.py}: Compute the mean average precision for each class against a test dataset.
\item \texttt{plot\_loss.py}: Plot the loss during training.
\item \texttt{yolov3\_detector.py}: The YOLOv3 detector implemented from scratch following this tutorial: \href{https://blog.paperspace.com/how-to-implement-a-yolo-object-detector-in-pytorch}{How to implement a YOLO object detector in pytorch}
\end{itemize}
The code for the dataset generation is in the subfolder \texttt{generate\_dataset/}.
Next to the dataset generation script itself, various helper scripts are included:
\begin{itemize}
\item \texttt{pyFrameexporter.py}: Export frames from videos
\item \texttt{pExportTransparentPNG.py}: Export cropped and masked PNG files of the individual objects from a uni-color background
\item \texttt{rename\_classes.py}: Rename classes within the labels of a dataset (used to combine multiple classes into one, or when the ID of an object changes)
\item \texttt{split\_test\_train.py}: Dividing a dataset in training and testing dataset
\item \texttt{remove\_outline.py}: A script to manipulate the outline of objects to add outlines, remove outlines and add glow around objects
\item \texttt{convert\_tensorflow\_to\_pytorch.py}: Convert Tensorflow label data to the format required to train YOLOv3
\item \texttt{check\_dataset\_integrity.py}: A script to check if a label exists for every image
\end{itemize}

%% Method
\section{Method}
\begin{figure}[t]
\centering
\includegraphics[width=0.4\textwidth]{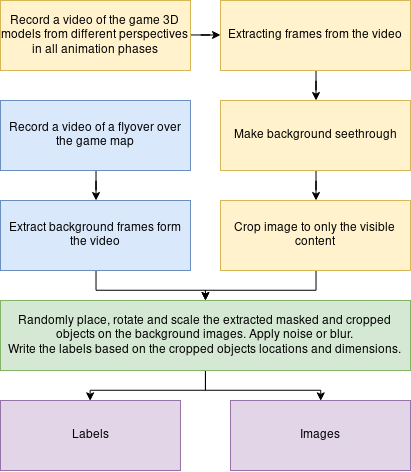}
\caption{Dataset generation pipeline.}
\label{fig:dataset_generation}
\end{figure}

The dataset generation method consists of 3 steps.
First the raw data is gathered form the 3D models.
In the second step, the raw data is masked and cropped to isolate the images of the objects.
In the third step the raw data is combined into randomized ``fake screenshots'', the synthetic training data.
A variety of parameters make it possible to add random variations such as noise, rotations and different objects to the images.
Furthermore, in this step the labels are created by using the dimensions of the masked and cropped object images to automatically generate the labels.
A detailed description of the dataset generation pipeline is shown in Figure \ref{fig:dataset_generation} and will be explained in this chapter.

\subsection{Generating raw data from 3D models}
First the raw data consisting of a set of masked images of the object in each animation phase and from different perspectives is generated.
\subsubsection{Viewing the 3D models}

\begin{figure}[t]
\centering
\includegraphics[width=0.4\textwidth]{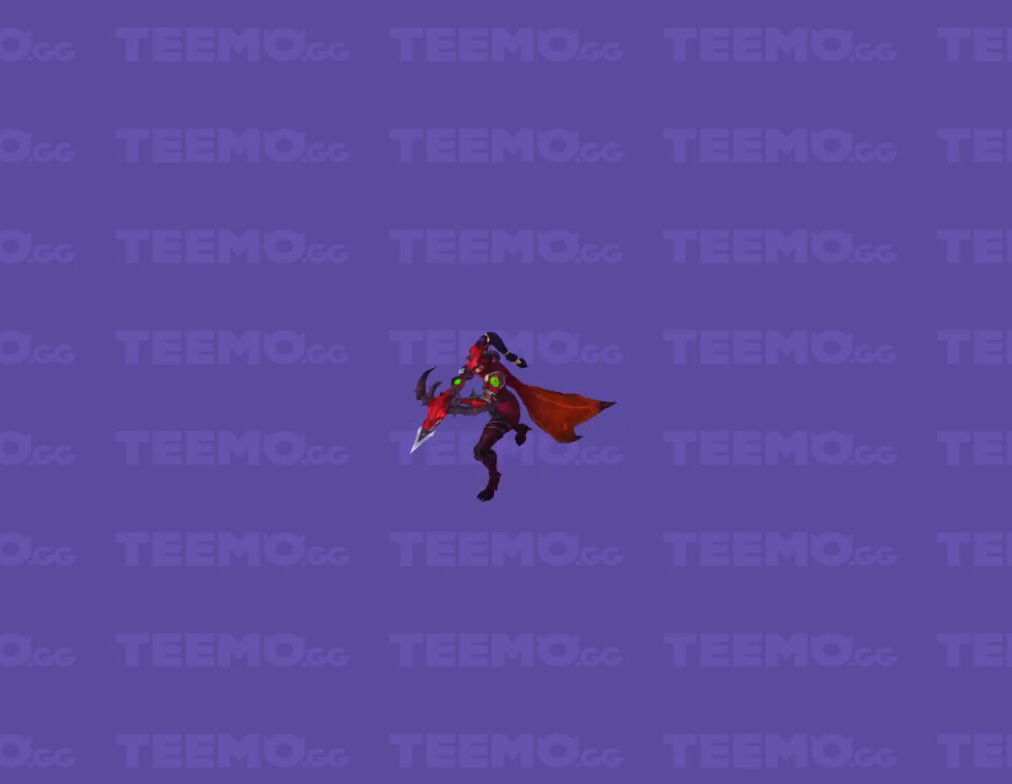}
\caption{3D model displayed in the model viewer.}
\label{fig:mv}
\end{figure}

The 3D models of the game characters are extracted and visualized using an online model viewer from \href{https://teemo.gg/model-viewer}{https://teemo.gg/model-viewer}.
The model viewer allows free rotation and zoom around the object as well as playing the characters animations.
Figure \ref{fig:mv} shows a character in the model viewer.
Another model viewer that can be run on a local machine is available on \href{https://github.com/Querijn/LeagueModel}{GitHub}.
\subsubsection{Recording a video and exporting frames}
A video of all animations and all perspectives is then recorded and for the current dataset every third frame is exported using the script \texttt{pyFrameexporter.py}.
Exporting every third frame created about 600-1000 raw object images which is enough to cover all possible animation phases and rotations of the objects.
Figure \ref{fig:mv} shows an exported frame with unicolor background.
The script has the following parameters:
\begin{itemize}
\item \textbf{frames:} Number of frames to skip between each exported frame
\item \textbf{resolution:} Output resolution of each frame
\item \textbf{prefix:} A text prefix to sort the output image files like for example ``output\_{}''.
\end{itemize}

\subsubsection{Masking and cropping the images}
\begin{figure}[t]
\centering
\includegraphics[width=0.2\textwidth]{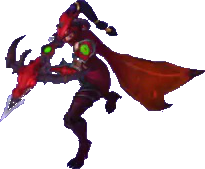}
\caption{Cropped and masked raw image.}
\label{fig:result}
\end{figure}

After extracting the images from the videos, the script \texttt{pyExportTransparentPNG.py} is used to remove the uni-color background by modifying the alpha channel of all uni-color pixels.
Afterwards the image is cropped to remaining visible content of the image.
The cropped dimensions of the image will alter be used as the bounding box of the image.
The result of the masking and cropping is shown in Figure \ref{fig:result}.

The script has the following parameters:
\begin{itemize}
\item \textbf{area:} Area of the input images on which the script is running. The rest of the image will be automatically removed. Especially if the position in the image and size of the object is known defining the area can help accelerate the script.
\item \textbf{background:} The color of the background that shall be transparent in the output image
\item \textbf{tolerance:} The tolerance for each RGB value. Helpful if the background is not exactly the same color everywhere or the outline of the objects in the output is not clean.
\item \textbf{remove\_outline:} Determines how many outline layers of the object should be removed.
Again, if the output objects have artifacts or bad outlines, this can help to remove them.
\end{itemize}

\subsection{Background image raw data}
The game background images are extracted from screenshots of the map recorded in the video game.
Other options would have been to use a high resolution image of the whole map and cut it into pieces.
Also the map is available as 3D model from: \href{https://sketchfab.com/3d-models/for-study-only-summoner-rift-3d-export-ac0a9c6676e34d1ebb184d8e93443c77}{Map 3D model}

\subsection{Other objects}
To further increase the robustness of the synthetic data, I also masked and cropped parts of the game's user interface and cursors.
These objects are placed on the screenshots as well to make the object detector learn what happens when a cursor is overlapping an objects and to not detect things on the UI.

\subsection{Generating synthetic training data}
With the raw data split in their categories (structures, minions, player characters), we can now generate synthetic game images.
Algorithm \ref{alg:1} and \ref{alg:2} describe the process.

\begin{algorithm}
\hspace*{\algorithmicindent} \textbf{Input:} dataset\_size, raw\_background, raw\_objects, num\_objects, max\_rotation, max\_scale\\
\hspace*{\algorithmicindent} \textbf{Output:} image, label
\begin{algorithmic}[1]
\caption{Generation of a synthetic labeled image}
\label{alg:1}
\Function{generate\_dataset}{}
\For{dataset\_size}
\State \textit{N} $\gets$ random([0,num\_objects])
\State cur\_image $\gets$ random(raw\_background))
	\For{\textit{N}}
	\State cur\_object $\gets$ random(raw\_objects)
	\State position $\gets$ random(cur\_image)
	\State scale $\gets$ random([-max\_scale, max\_scale])
	\State rotation $\gets$ random([-max\_rotation, max\_rotation])
	\State \texttt{add\_object(cur\_image, cur\_object, rotation, scale, object\_class, position)}
	\EndFor
\State \texttt{apply\_noise()}
\State \texttt{apply\_blur()}
\EndFor
\EndFunction
\end{algorithmic}
\end{algorithm}

\begin{algorithm}
\hspace*{\algorithmicindent} \textbf{Input:} cur\_image, raw\_object, rotation, scale, object\_class, position
\\
\hspace*{\algorithmicindent} \textbf{Output:} image, label
\begin{algorithmic}[1]
\caption{Add an object to an image}
\label{alg:2}
\Function{add\_object}{}
\State \texttt{scale\_object(scale)}
\State \texttt{rotate\_object(roation)}
\For{\textbf{all} cur\_pixels \textbf{in} cur\_image}
	\If {cur\_pixel is covered by object}
	\State cur\_pixel $\gets$ object\_pixel
	\Else
	\State cur\_pixel $\gets$ cur\_pixel
	\EndIf
\EndFor
\State \texttt{append\_to\_labels(obejct\_class, object\_center, width, height)}
\EndFunction
\end{algorithmic}
\end{algorithm}

\begin{figure*}[t]
\centering
\includegraphics[width=0.8\textwidth]{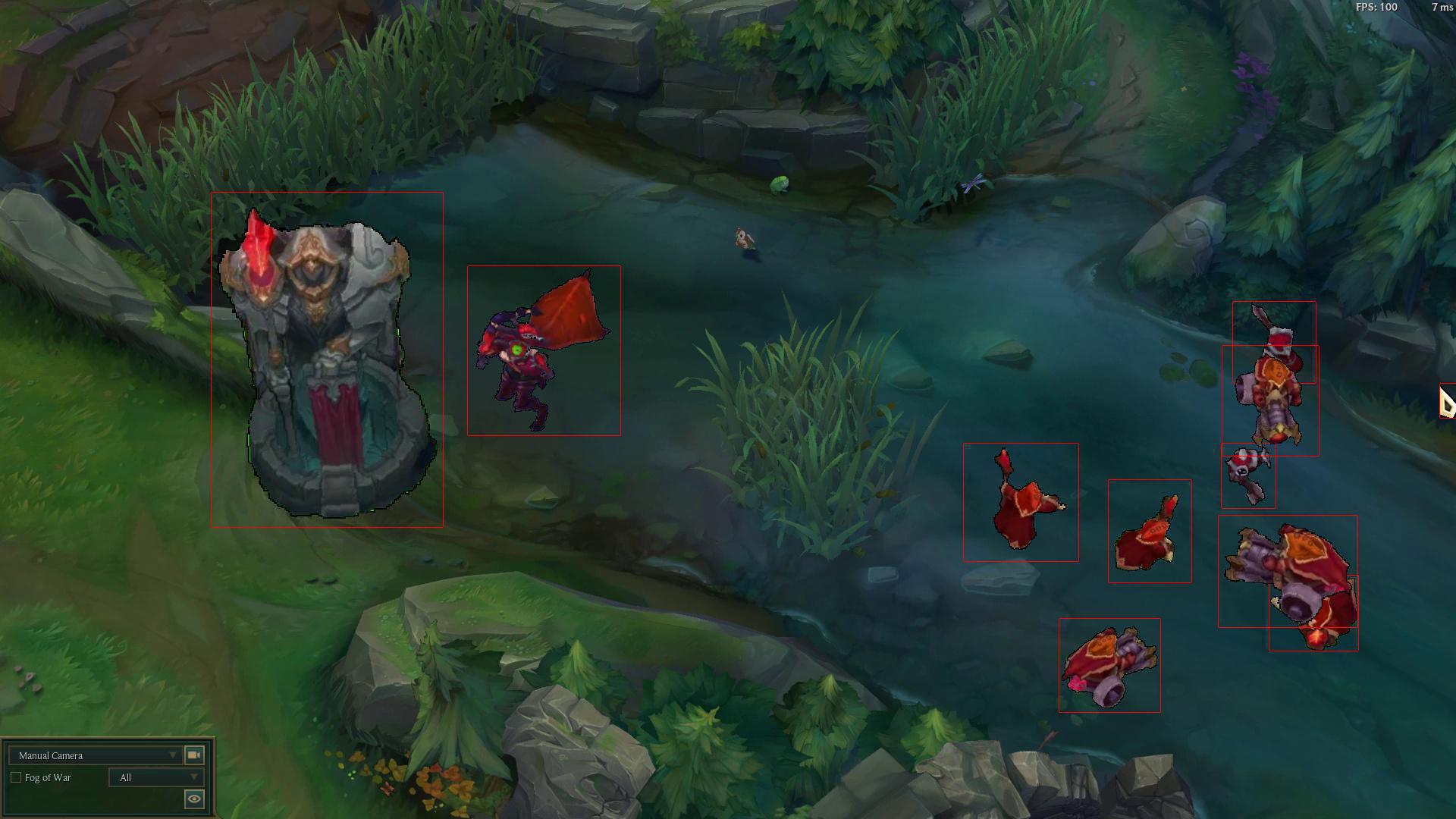}
\caption{An automatically generated labeled image. The red boxes visualize the generated label. Note how the minions are overlapping and clustered around the bias point to simulate their behavior of clumping up during fighting.}
\label{fig:example}
\end{figure*}

Figure \ref{fig:example} shows an artificially generated labeled data. For the sake of visualization no noise or blur have been applied. 

In the following sections different variations and parameters of the algorithm are explained.
Note that each of the additional ways of modifying the images can be turned off individually.

\subsubsection{Placement}
In this step the raw data is placed on a randomly selected background image.
The placement of the objects is important to generate realistic game screenshots and to make the object detector more robust.
During the game, the objects will overlap.
On top of the game the cursor and the user interface will overlap the game.
To reproduce this layering, the following parameters allow to modify the way in which the objects are placed.
\begin{itemize}
\item \textbf{cursor\_max/min, champion\_max/min, minion\_max/min, tower\_max/min}: Set the minimum/maximum number to control how often objects are placed. A number of objects between the min and max value will be uniformly picked.
\item \textbf{overlay\_chance}: Probability of adding UI elements to the image 
\item \textbf{fog\_of\_war}: Probability to add fog of war to the image
\end{itemize}

The position at which an object is placed is chosen randomly by uniformly selecting a \textit{x} and \textit{y} coordinate in the image.

\paragraph*{\textbf{Grouping minions}}
In the real game groups of computer controlled minions will usually fight each other and thus usually group up and overlap more often.
To simulate this effect, the minions are not only placed randomly on the image, but can be placed around a bias point using a normal distribution.
The parameter \textbf{bias\_strength} can be used to adjust the way the minions are clustered.

\subsubsection{Scale and rotate}
Parameters allow to set the input scale and orientation of each object in the image.
Additionally, the parameters \textit{max\_scale} and \textit{max\_rotation} set the maximum amount of the random scaling and rotation applied to the input values.ö
Choosing scale and rotation for an object is picking a rotation and scale value between the given maximum values and add/subtract it from the input orientation and scale.
The parameter \textbf{sampling\_method} can be used to chose different sampling methods for scaling the image.
Since the results of different sampling methods can look different, this can further improve the robustness of the dataset.

\subsubsection{Noise and blur}
To make the dataset more robust to in game effects like particles covering or coloring the characters, noise is introduced.
Random noise is applied by randomly chaning the RGB values of the final image.
The parameter \textbf{noise} can be used to adjust the strength of the noise for each individual channel.
Gaussian blur is applied to smoothen the image.
The parameter \textbf{blur\_strength} is used to set the strength of the blur.

\subsubsection{User Interface}
During testing the user interface of the real game was not included in the training data and thus led to significant wrong detections.
Therefore, I created masked versions of the game UI and replace champion specific icons with a random selection of icons.
Another problem was the quite frequently occurring overlapping of the player's mouse cursor especially with minions (in order to attack them for example).
To solve this I created raw images of the cursor graphics which are added in random locations on the image.

%% Empirical Evaluation
\section{Implementation and evaluation of the object detector}

In this section I will describe the YOLOv3 structure of the convolutional neural network used to train the synthetic model $\mathcal{S}$ and the hand labeled model $\mathcal{H}$.
The datasets used are described and the training parameters for each of the models are explained.
Finally I will explain the method of comparing the two models and discuss the results.

\subsection{YOLOv3 model}
\begin{figure}
\centering
\includegraphics[width=0.4\textwidth]{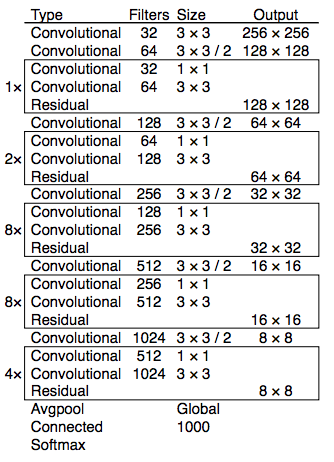}
\caption{Structure of the YOLOv3 Darknet-53 neural network \cite{redmon2018yolov3}.}
\label{fig:darknet}
\end{figure}

The neural network structure as introduced in \cite{redmon2018yolov3} used for feature extraction is called \textit{Darknet-53} and shown in Figure \ref{fig:darknet}.
The network is a fully connected convolutional network and has 75 convolutional layers overall.
The network consists of multiple blocks marked by boxes in Figure \ref{fig:darknet}.
Each block has an additional shortcut connection, similar to skip connections used in ResNet, around it.
The blocks consist of two successive convolutional layers with kernel size $3$ and kernel size $1$ followed by a $Residual$ layer.
In between each block a convolutional layer with stride 2 is used to downsample the feature maps.
Also no pooling is used to prevent the loss of low-level features.
Lastly global average pooling is applied.
The output layer is a fully connected layer with 1000 nodes followed by a softmax activation function.

Each unit in the resulting feature map can predict up to 3 bounding boxes.
A bounding box consists of the box's center coordinates, its width, height, the objectness score and the class confidence.
The network predicts 4 coordinates for each bounding box: $t_x, t_y, t_w, t_h$.
Together with the bounding box offset within the image $c_x, c_y$ and the box prior $p_w, p_h$ the bounding box center coordinates and dimensions are computed as
\begin{equation}
\begin{split}
b_x = \sigma(t_x) + c_x\\
b_y = \sigma(t_y) + c_y\\
b_w = p_we^{t_w}\\
b_h = p_he^{t_h}\\
\end{split}
\end{equation}
During the training the loss is computed as the sum of squared errors.

More details on the prediction of the bounding boxes, the objectness score and the class confidences can be found in \cite{redmon2018yolov3}.

\subsection{The training data}
Information about the training datasets can be seen in Table \ref{tab:1}.
The synthetic dataset consisted of 6200 automatically generated imagtes with different parameter variations.
All images were in JPEG format with their corresponding labels being txt files in the format:\\
\textit{<object class> <x center> <y center> <width> <height>}.

The old dataset consisting of 481 images was annotated using the tool \href{https://github.com/tzutalin/labelImg}{LabelImg}.
The tool generate xml labels with a different format, thus I wrote a script to convert the xml labels to the above format.

The difference between the two datasets is that the synthetic data can distinguish the different types of minions and therefore has 5 detectable object classes instead of 3.
The following classes are included in the synthetic dataset:
\begin{itemize}
\item Red side towers
\item Red melee minion
\item Red caster minion
\item Red canon minion
\item Champion Vayne (red skin, because back then when I labeled the old dataset I thought this would somehow to contrast from the background, but I also made a synthetic dataset with the default skin)
\end{itemize}
In the hand labeled dataset the minions are grouped into one class.

In case one wants to train the synthetic model with only three classes, combining the minions into one class, I wrote a script that can do so by replacing object class IDs: (\texttt{rename\_classes.py}).

\subsection{Training the models}
The models have been trained using \textit{darknet}~\cite{darknet13}\footnote{\href{https://pjreddie.com/darknet/}{Darknet Website}, \href{https://github.com/pjreddie/darknet}{GitHub}}, an open source neural network framework written in C.
I selected this method because it was easy to setup and run on GPUs. Since I implemented the network and the detection code myself I could have written a training script myself, but for saving time on this already huge project I resorted to use this implementation.

Note that the new model trained on the synthetic dataset is additionally capable of differentiating the 3 different minions classes instead of grouping them as just enemies.
Therefore, $\mathcal{S}$ is capable of detecting 5 classes compared to $\mathcal{H}$ which can only detect 3.
This means of course that we require more data to train $\mathcal{S}$ and a longer time to train.

\subsubsection{Training on hand labeled data}
The parameters used when training $\mathcal{S}$ can be seen in Table \ref{tab:1}.
The model was trained on one Nvidia 1080Ti GPU with a batch size of 64. The image width differed, because the old labeled training data I had was on images with a resolution or 1280x1024 compared to the new images with 1920x1080.
However, this should not be a problem since the network used is a fully connected convolutional network and the input size of the images does not matter as long as it stays constant.
The angle parameter sets the maximum random rotation of the images while training.
Since the objects to train are already rotated in the dataset generation and the game is always played from the same perspective, this is not necessary.
The Saturation, Exposure and Hue values set the random values for further random image manipulation.
The values use are the standard values.
The learning rate was also left at standard values.
The burn-in episodes were set to 500.
The burn in time is the number of episodes in the beginning of the training during which the learning rate will grow from 0 to its intended value.
I have read online that this has been found to increase the training speed, but I have not found any papers that show that and I have been able to verify this myself.
The model was trained for 7500 episodes, because the relatively low size of the training dataset could lead to overfitting if more episodes were used.

\subsubsection{Training on synthetic data}
\begin{figure}[t]
\includegraphics[width=0.5\textwidth]{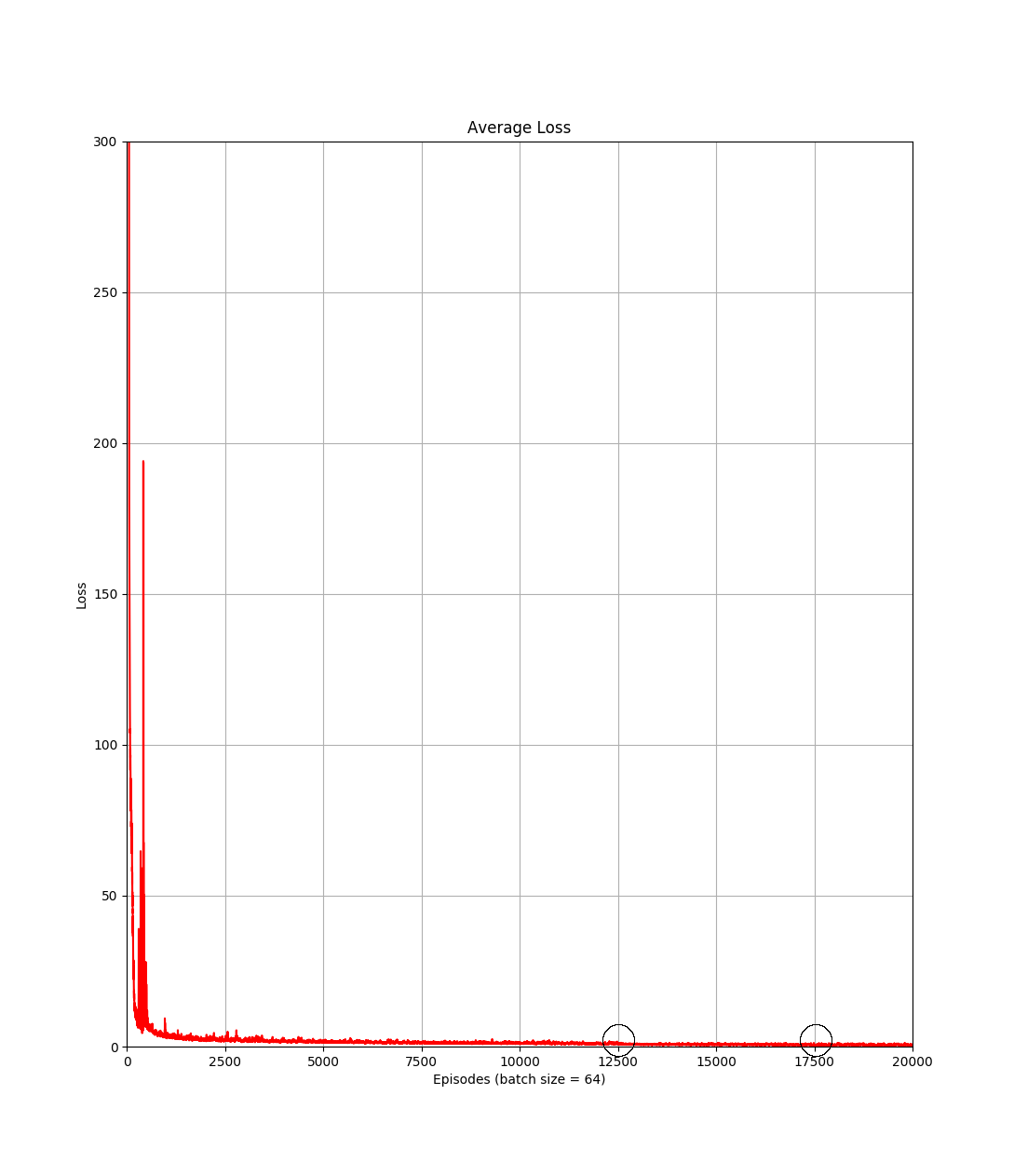}
\caption{Loss per episode during training of the model with the synthetic dataset. Training for 40000 episodes with batch size 64 on two Nvidia 1080Ti GPUs. The scale is only until 20000 episodes, because when using two GPUs darknet halfs the number of episodes.}
\label{fig:loss}
\end{figure}
\begin{table*}[h]
\centering
\caption{Training config parameters for synthetic dataset $\mathcal{S}$ and hand labeled dataset $\mathcal{H}$.}
\label{tab:1}
\begin{tabular}{|l|ccc|}
\hline
	Model 			& $\mathcal{S}$ & $\mathcal{H}$ & $\mathcal{C}$\\
\hline
	Number of images& $6200$		& $481$	& $6681$\\
	Image resolution& $1920 \times 1080$	& $1280 \times 1024$ & $1920 \times 1080 $ and $1280 \times 1024$ \\
	Classes			& $5$			& $3$ & 3 \\
	Dataset creation time & 30 min & 24 hours & -\\
\hline
	Batch size		& $64$			& $64$  & $64$ \\
	Image width 	& $960$ 		& $640$	& $960, 640$\\
	Angle			& $0$ 			& $0$ 	& $0$\\ 
	Saturation		& $1.5$ 		& $1.5$ & $1.5$ \\
	Exposure		& $1.5$ 		& $1.5$ & $1.5$ \\
	Hue				& $0.1$ 		& $0.1$ & $0.1$ \\
\hline
	Learning Rate	& $0.001$ 		& $0.001$ & $0.001$	\\
	Burn In	Episodes &$1000$		& $500$   & $1000$	\\
	Max. Episodes	& $40000$ 		& $7500$  & $40000$	\\
	Reduce LR at	& $25000, 35000$& $5000$  & $25000, 35000$	\\
	Reduce LR by	& $0.1, 0.1$ 	& $0.1$	 & $0.1, 0.1$	\\
\hline
	Training time   & $70$ hours (2 GPUs) & $12$ hours (1 GPU) & $70$ hours (2 GPUs) \\
\hline
\end{tabular}
\end{table*}

The parameters for the synthetic data can be seen in Table \ref{tab:1}.
The batch size of 64 was effectively twice as large as for training $\mathcal{H}$ because the training ran on 2 GPUs. Therefore the plot in Figure \ref{fig:loss} is from $0$ to $20000$ with different learning rate reduction values.
Similar to the training of $\mathcal{H}$, the angle, saturation, exposure and hue values are unchanged.
The training loss on the synthetic dataset is shown in Figure \ref{fig:loss}.
It can be seen that the loss declined quite rapidly in the first 2500 episodes.
However, when testing models trained for less than 10000 episodes, the mAP of $\mathcal{S}$ was significantly lower than the mAP of $\mathcal{H}$ trained with only 7500 episodes.
This is a result of the fact that $\mathcal{S}$ was trained with more images and with more different classes.
The learning rate for $\mathcal{S}$ was the same as the standard value used for existing YOLOv3 models.
The burn-in rate here was selected larger because more maximum episodes were used.
Furthermore the reduction of the learning rate (marked as black circles), helped the learning process of $\mathcal{S}$ as can be seen from the decrease of the loss around 25000 episodes.
The reduced learning rate was introduced to prevent overfitting of the model.

\subsection{Comparing the datasets}
\begin{figure}[h]
\centering
\includegraphics[width=0.2\textwidth]{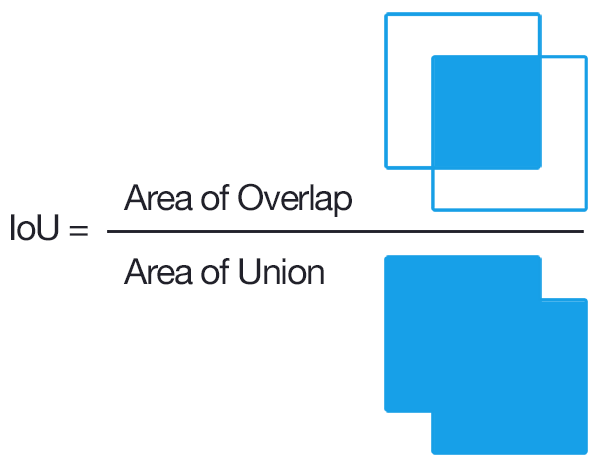}
\caption{Computation of the Intersection to Union ratio. Image from \href{https://www.pyimagesearch.com/wp-content/uploads/2016/09/iou_equation.png}{Link}.}
\label{fig:iou}
\end{figure}

\begin{table*}
\centering
\caption{Mean average precision on the test set per class and model. Tracking of the player character in percent of the whole video.}
\label{tab:2}
\begin{tabular}{|l|ccc|}
\hline
	Model 			& $\mathcal{S}$ & $\mathcal{H}$ & $\mathcal{C}$ \\
\hline
	Input resolution& $640 \times 360$ & $640 \times 524$ & $960 \times 540$\\
\hline
	mAP Tower		& $0.81$	& $0.5$  & $0.75$ \\
	mAP All minions & $0.85$ (weighted average)	& $0.83$ & $0.83$ \\
	mAP Canon		& $1.0$		&		 & \\
	mAP	Caster		& $0.93$	& 		 & \\
	mAP Melee		& $0.73$	&		 &\\
	mAP Vayne		& $1.0$		& $0.81$	& $0.94$\\
\hline
	Character tracking time in percent of the whole video: & & &\\
\hline
	Correctly detected character & $88.53\%$ & $84.12\%$ & $73.2\%$ \\
	Detected character multiple times & $8.87\%$ & $0.28\%$ & $1.73\%$ \\
	No detection 	& $2.6\%$ & $15.6\%$ & $25.1\%$\\
	FPS	on a GTX1080		& $18$		& $12$ & $10$\\
\hline
\end{tabular}
\end{table*}
In order to compare the models, I created a hand labeled test set of 54 images.
On this test set I computed the mean average precision (mAP) as defined in the \href{http://host.robots.ox.ac.uk/pascal/VOC/voc2012/}{PASCAL VOC 2012 competition (Link)} using the script \texttt{LeagueAI\_mAP.py}.
The mAP is computed as the number of correct classifications of an object (true positives) by the number of occurences in the dataset.
An object counts as correctly classified if their Intersections divided by their union area \textit{IoU} >= 0.5.
The computation of the IoU of an object is visualized in Figure \ref{fig:iou}.
For each object class $I$ I counted the correct ($IoU >= 0.5$), wrong (different class or $IoU < 0.5$) and not detected ground truth labels and divided them by the total number of class occurrences in the dataset $T$.

\begin{equation}
mAP = \frac{1}{T} \sum^{I}_{i=0} 1~|~\{IoU_i >= 0.5\}
\end{equation}

To evaluate the performance of both model in a real-time environment I used the video from which the hand labeled test dataset was generated.
The 126 seconds video clip was recorded in a normal game setting and includes tests such as zooming in, using abilities and attack minions.
In order to play the game it is crucial to know the location of one's player character.
Therefore I computed how long each model was able to detect the player character.

\subsection{Results}
\begin{figure}[H]
\includegraphics[width=0.5\textwidth]{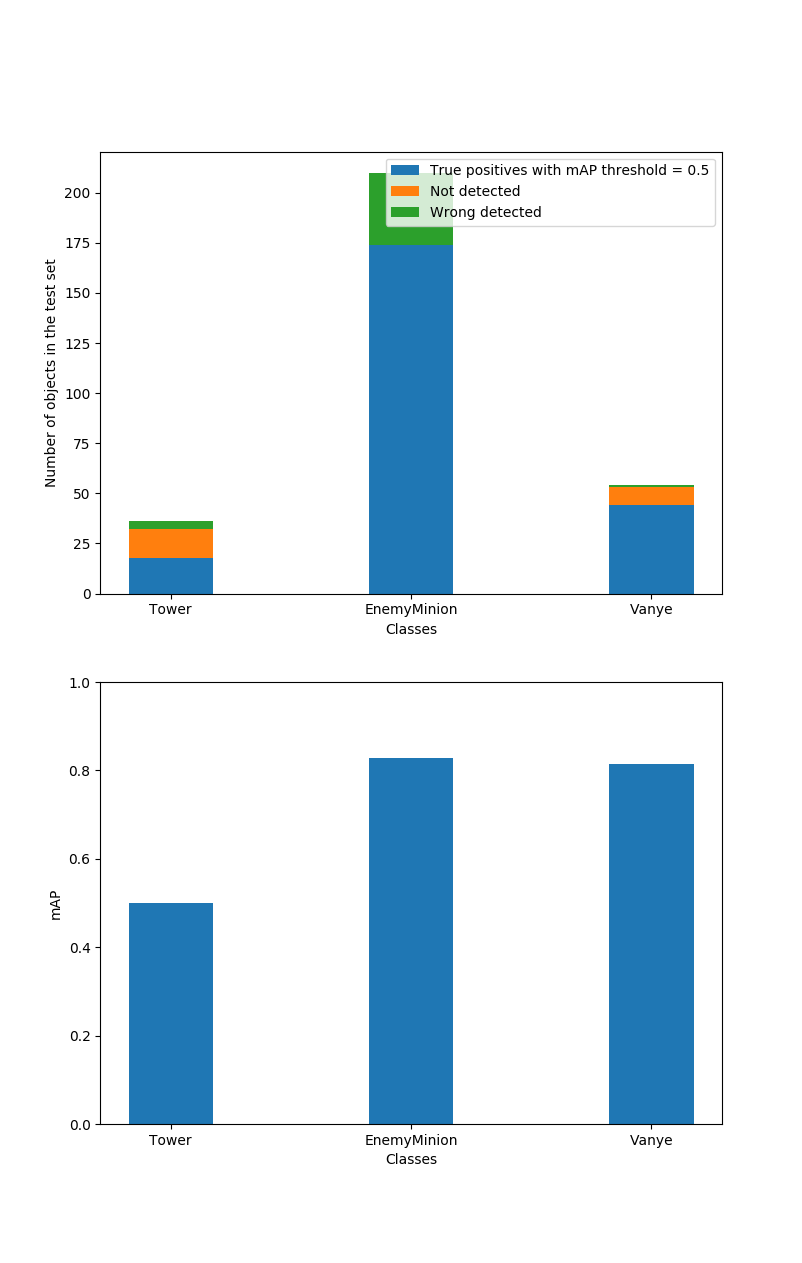}
\caption{Mean average detection precision per class on a test set of 54 hand labeled images for the old model tained on hand labeled data.}
\label{fig:results_old}
\end{figure}

\begin{figure}[H]
\includegraphics[width=0.5\textwidth]{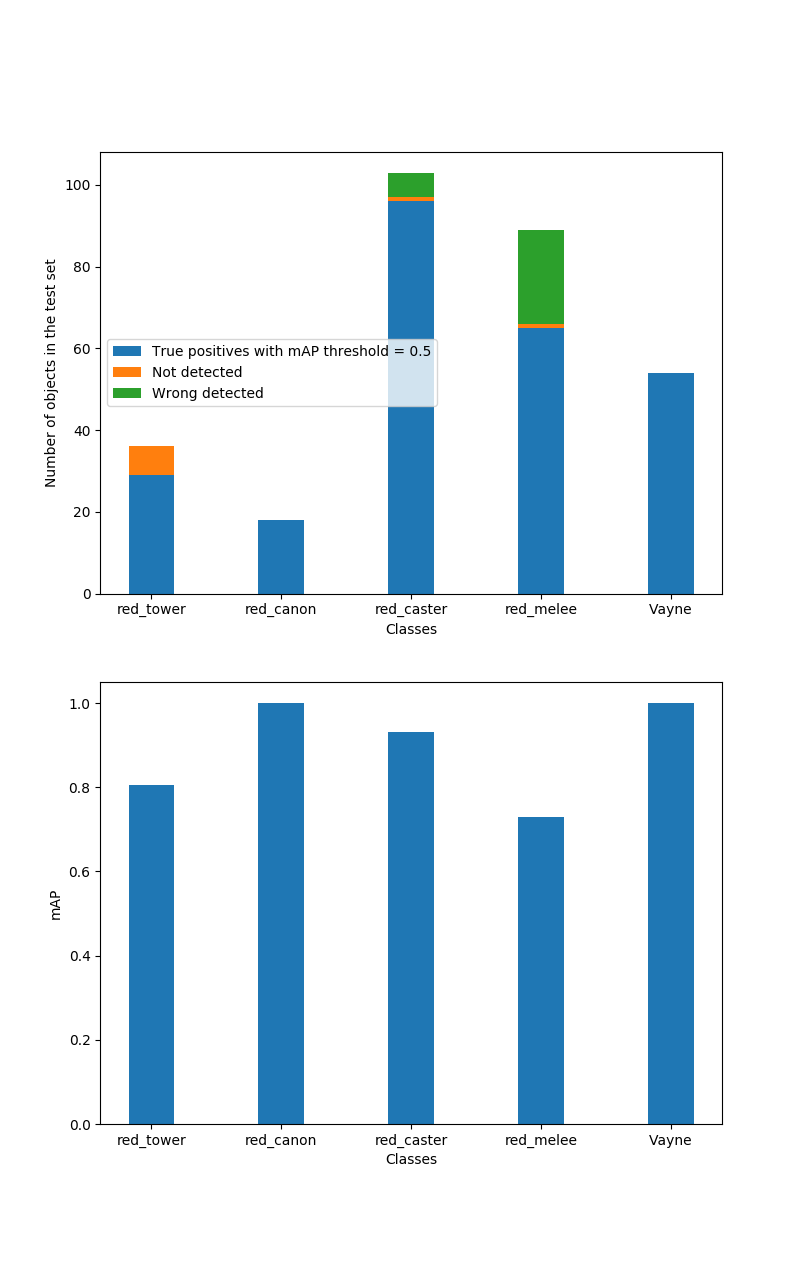}
\caption{Mean average detection precision per class on a test set of 54 hand labeled images for the new model trained on synthetic data. Note that here the minion category is split up in the three different types of minions.}
\label{fig:results_new}
\end{figure}

The results of both models on the test set are shown in Figure \ref{fig:results_old} for $\mathcal{H}$ and \ref{fig:results_new} for $\mathcal{S}$.
Table \ref{tab:2} shows the numerical results from comaring the mAP of all three models on a hand labeled test dataset and a video.
The combined dataset $\mathcal{C}$ will be discussed later as there might have been problems with the training and the results do not allow for an accurate comparison.

Also important to note is that the comparison is in favor of the hand labeled model, since it only has to differentiate between 3 classes while the synthetic model is also capable of telling apart the different types of minions.
The champion Vayne and the Tower class however are directly comparable.
Furthermore, testing showed that while mode $\mathcal{S}$ was trained on higher resolution data, it performed better for lower resolution input images leading to a higher mAP and more frames per second processed.
The same effect could not be confirmed on the hand labeled data.
The input resolution of the test data is shown in Table \ref{tab:2}.
Due to the lager dataset and more object classes, training model $\mathcal{S}$ took much longer at 70 hours on two Nvidia GTX 1080Ti GPUs compared to 12 hours on one GPU for model $\mathcal{H}$.

The results of the mAP on the test dataset showed that in the test set most of the objects were minions (which is usually in the game the case since they appear in groups of 6-7).
Also the canon minion type is more rare.
Furthermore, we can see that most wrong detections happened among the melee and caster minions.
This is because they usually move in groups of 3 each and overlap each other during fighting.
The worst performance was recorded for the melee minion, resulting from the overlapping with another additional group of minions currently not detected by the model: the melee minions of the player's team.

Comparing the mAP on the hand labeled test dataset shows that model $\mathcal{S}$ provides a higher mean average precision across all classes while at the same time being able to discern more classes.
The relatively low performance of both models on the tower, a static object, is most likely a result of the size of the tower.
Compared to the other game objects it is larger and therefore most of the time only partially on the screen and often overlapped by other objects.
The findings in this report do not confirm the findings in \cite{rajpura2017object}, that the synthetic data leads to less precise object detection.
However, the 2D nature of the data generation on a video game with a limited number of objects and animation phases is not as complex as a 3D environments.
Especially considering that video games are usually designed in a way that the objects are easily discernible by the human players and therefore probably also for an object detector.

To evaluate the performance in a real-time scenario, I used the video from which the 54 test images were extracted and computed the average tracking times of the player character.
The results are listed in Table \ref{tab:2}.
It can be seen that both models could track the player character more than $84\%$ of the time where mode $\mathcal{S}$ has a slight advantage with $88.5\%$.

Interestingly, model $\mathcal{H}$ was 7 times more likely to not detect the character than $\mathcal{S}$.
On the other hand $\mathcal{S}$ 40 times more likely to detect the player character and other objects that were mistaken for the player character.
However, multiple detections are not as big of a problem as losing tracking entirely since sorting the list of possible character detections by their confidence usually gives the correct object as the detection with the highest confidence.
Also noteworthy is that as a result of the lower image resolution, the average FPS using model $\mathcal{S}$ was 50\% higher.
Therefore we can conclude that in a real time scenario the synthetic data leads to a more precise model than hand labeled data.

\begin{figure*}
\centering
\includegraphics[width=\textwidth]{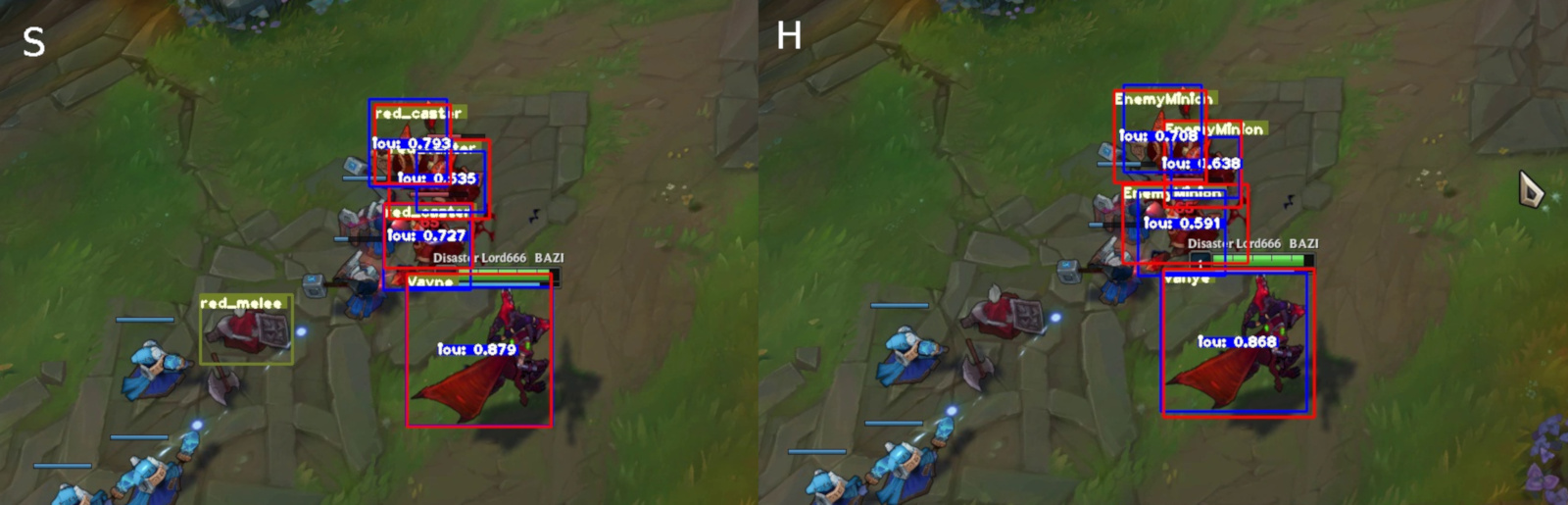}
\caption{Model $\mathcal{S}$ detects dead minions. While this shows the better performance of $\mathcal{S}$, this is a problem that needs to be addressed in the future.}
\label{fig:new_wrong}
\end{figure*}

Lastly I will analyze three specific cases from the 54 test images that show the difference between the two models.
Figure \ref{fig:new_wrong} shows the result of the mAP computation for model $\mathcal{S}$ on the left and model $\mathcal{H}$ on the right.
The predictions of the object detector are marked in red and the ground truth (hand labeled) is marked in blue.
For each detected object that has a ground truth counterpart the IoU ratio is shown.

In Figure \ref{fig:new_wrong} both models detected all objects correctly, we can see that $\mathcal{S}$ detected in addition a melee minion that had already died.
Dead minions were not part of the synthetic training data, indicating that the model is capable of generalizing to unseen data.
The synthetic data takes more perspectives of all the objects and introduces noise and is thus able to better generalize to unseen objects.
On the other hand detecting dead minions is not a desired behavior of the object detector and thus in the future it could help to add dead minions as negative examples to the training dataset in order to prevent these missdetections.
\begin{figure*}
\centering
\includegraphics[width=\textwidth]{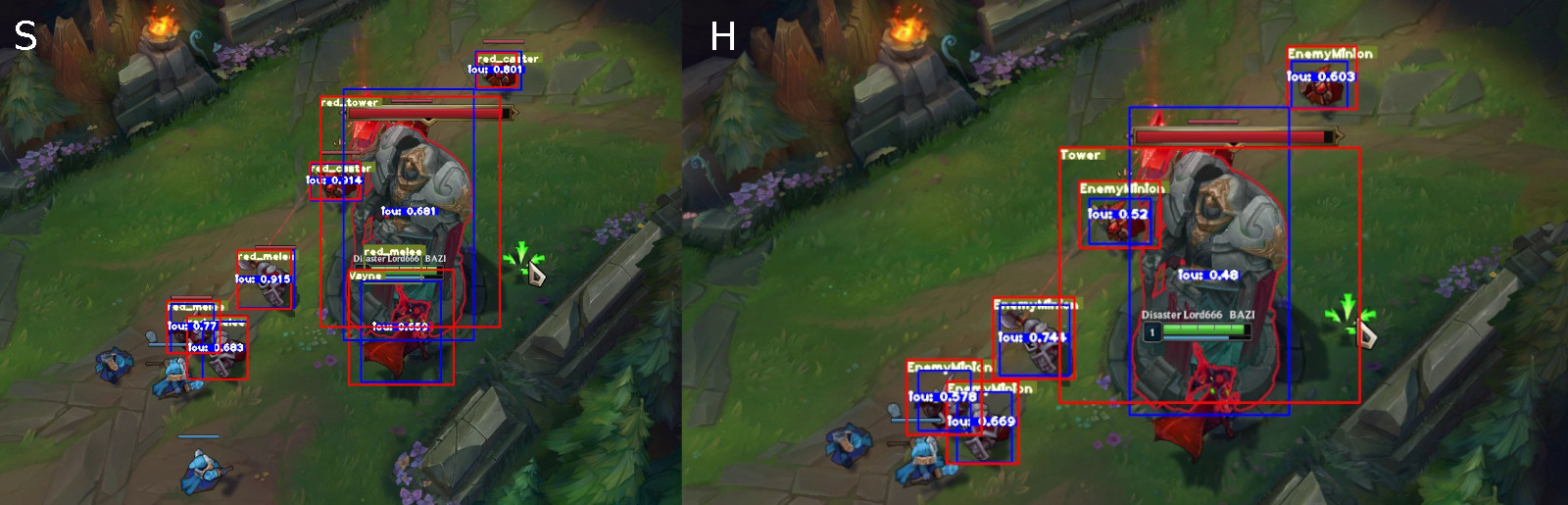}
\caption{Model $\mathcal{S}$ is capable of handling overlapping objects better.}
\label{fig:new_better}
\end{figure*}

Figure \ref{fig:new_better} shows an example where model $\mathcal{S}$ clearly outperforms $\mathcal{H}$.
We can see that on the left side the predicted bounding boxes match the ground truth better and the IoU ratios are higher.
Furthermore, model $\mathcal{H}$ was not able to detect the player character which was overlapping the tower.
In the game this would lead to a dangerous loss of tracking next to a dangerous map objective such as an enemy tower.
We can see that $\mathcal{S}$ is better able to handle overlapping objects.
\begin{figure*}
\centering
\includegraphics[width=\textwidth]{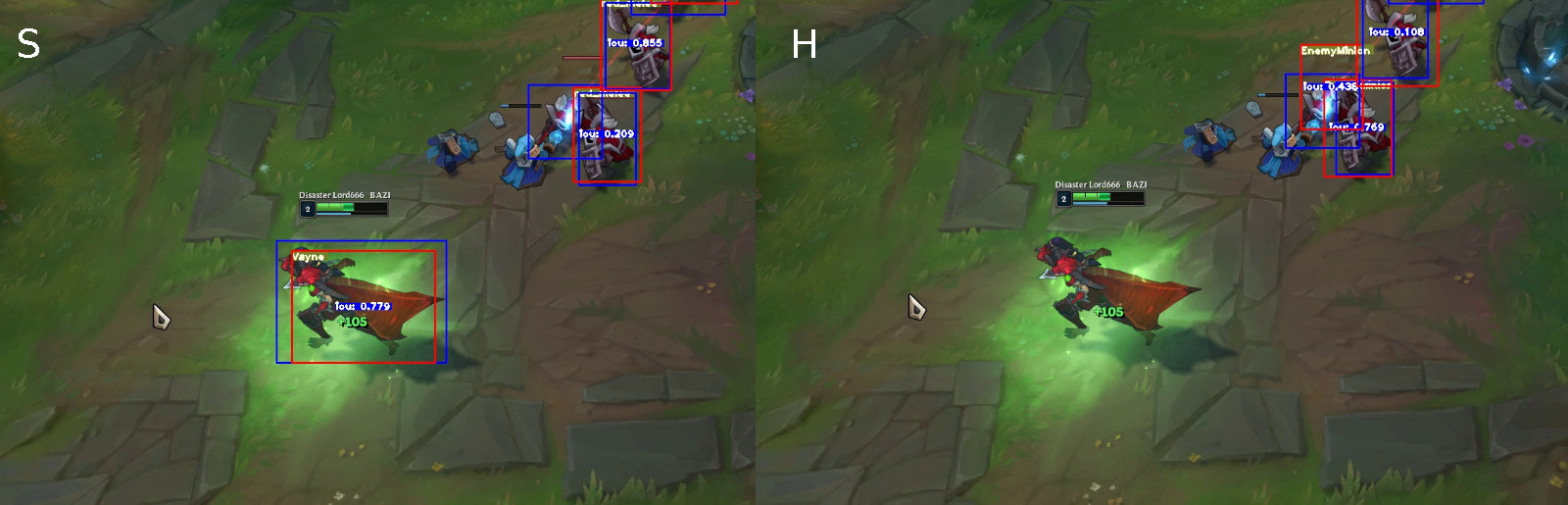}
\caption{Model $\mathcal{S}$ is better able to generalize and can still detect the player character while using abilities that add additional particles.}
\label{fig:new_better2}
\end{figure*}
In Figure \ref{fig:new_better2} we can see another example for the increased flexibility of model $\mathcal{S}$.
In this situation the player character used a spell which leads to the display of additional green particles around and on top of the character.
We can see that $\mathcal{S}$ was still capable of detecting the champion.
This is most likely a result of the introduced blur and noise as well as the larger amount of training data, allowing $\mathcal{S}$ to better handle the additional color changes.

\subsection{Combining the datasets}
In order to confirm the findings in \cite{rajpura2017object} and \cite{prakash2018structured}, that combining hand labeled and synthetic data can improve the performance of object detectors, both the datasets from $\mathcal{S}$ and $\mathcal{H}$ have been combined and used to train a model $\mathcal{C}$.
As can be seen in Table \ref{tab:2} the performance of $\mathcal{C}$ was in between $\mathcal{S}$ and $\mathcal{H}$.
In the video comparison the combined model actually performs worse than the other models, with most of the errors being cause by not detecting the player character.
This does not match the expectations as more different hand labeled data should at least provide similar performance to the synthetic data.
The reason for this is most likely the different aspect ratio of the synthetic and hand labeled images.
Furthermore, I had to reduce the number of classes in the synthetic dataset to 3 by combining all minion types into one.
Therefore, retraining the combined model with similar data is very likely going to further improve the performance.
In the future I will probably create a new hand labeled dataset for the minions of both teams, because Figure \ref{fig:results_old} and \ref{fig:results_new} as well as Table \ref{tab:2} show that the detection on the minions is the most unprecise and could therefore benefit from additional human labeled data.
Additionally hand labeled data on these smaller, often overlapping and more frequent occurring objects might improve detection quality.

%% Conclusion and Future works
\section{Conclusion and future works}
In this technical report I presented my method of training an object detector for a video game using synthetically generate training data.
The performance of a model trained on the synthetic data was compared to a model trained on hand labeled data.
I trained three models and compared their performance by computing the mAP on a set of hand labeled test images and by comparing their tracking time of an object in a video clip of the game.
The model trained on the synthetic data was capable of reaching higher mAP on more classes (5 instead of 3) and was able to more consistently track the player character in a real-time scenario.
Furthermore, the time spent to generate the training data was significantly lower for the automatically generated training data.
Analyzing the individual detections shows that model $\mathcal{S}$ is generally better able to generalize to unseen data and handle overlaps but sometimes detects objects it should not detect them (dead minions).
This indicates that by combining hand labeled and synthetic data, the detection performance could be further improved.
Especially on classes that overlap often and appear most of the time, like minions, adding hand labeled data could significantly improve detection precision.

The results of previous research in \cite{rajpura2017object} and \cite{prakash2018structured}, that combining synthetic and hand labeled data would further improve the performance of the object detector, could not be confirmed.
The performance of the combined dataset was better than the hand labeled data but worse than the synthetic data.
The different aspect ratios of the old hand labeled dataset and the new synthetic images are likely the reason for the lower performance.
Since the project has moved to higher resolution training data and already shows good performance, repeating the performance evaluation with new hand labeled data is not a high priority.
For the future, I plan to develop a method to fuse automatic and hand labeling capabilities especially for smaller objects that occlude each other often like the minions.
Especially in these cases the detection accuracy could be improved as suggested by the related literature.

In future work the dataset generation will be expanded to a many more object classes (possibly all 140+ champions as well as the blue teams minions and structures).

Lastly, the generation of raw data to generate the training data could be further improved.
I found out that a 3D model of the game map is available for download.
Thus instead of using video recordings of the model viewer and merging it with images of the game map, I would directly use the map 3D model and place the 3D objects directly into the map to generate even more realistic scenes.
However, considering the already quite good performance and the amount of work required for this, I will not prioritize this approach.

The next steps for the LeagueAI framework will be to extract information about the health/mana of the characters and their position locally in the screen and globally on the map.

%%% Bibliography
\bibliographystyle{ieeetr}
\bibliography{report}%

\end{document}